\documentclass[11pt]{article}

\usepackage[final]{acl}

\usepackage{times}
\usepackage{latexsym}

\usepackage[T1]{fontenc}

\usepackage[utf8]{inputenc}

\usepackage{microtype}

\usepackage{inconsolata}

\usepackage{graphicx}

\usepackage{afterpage}
\usepackage{amsmath}

\usepackage{listings}
\usepackage{xcolor}

\usepackage{tabularx}
\usepackage{array}
\newcolumntype{C}{>{\centering\arraybackslash}c}
\usepackage{makecell}
\usepackage{diagbox}
\usepackage{tikz}

\usepackage{booktabs}
\usepackage{subcaption}
\usepackage{multirow}

\usepackage{float}

\usepackage{lipsum}
\usepackage{threeparttable}
\usepackage{pifont}

\usepackage{caption}
\title{Explainable Multimodal Aspect-Based Sentiment Analysis with Dependency-guided Large Language Model}

\author{
\vspace{0.1cm}
Zhongzheng Wang\textsuperscript{\ding{169}\ding{168}} \quad
Yuanhe Tian\textsuperscript{\ding{170}\ding{168}}\thanks{Corresponding author.} \quad
Hongzhi Wang\textsuperscript{\ding{169}} \quad
Yan Song\textsuperscript{\ding{171}} \\
\textsuperscript{\ding{169}}Harbin Institute of Technology \quad
\textsuperscript{\ding{168}}Zhongguancun Academy \\
\textsuperscript{\ding{170}}Zhongguancun Institute of Artificial Intelligence \\
\vspace{0.1cm}
\textsuperscript{\ding{171}}University of Science and Technology of China \\
\textsuperscript{\ding{169}}25b903046@stu.hit.edu.cn\quad
\textsuperscript{\ding{170}}yhtian94@gmail.com \\
\textsuperscript{\ding{169}}Wangzh@hit.edu.cn \quad
\textsuperscript{\ding{171}}clksong@gmail.com
}

\begin{document}

\maketitle

\begin{abstract}

Multimodal aspect-based sentiment analysis (MABSA) aims to identify aspect-level sentiments by jointly modeling textual and visual information, which is essential for fine-grained opinion understanding in social media.
Existing approaches mainly rely on discriminative classification with complex multimodal fusion, yet they lack explicit sentiment explainability.
In this paper, we reformulate MABSA as a generative and explainable task, proposing a unified framework that simultaneously predicts aspect-level sentiment and generates natural language explanations.
Based on multimodal large language models (MLLMs), our approach employs a prompt-based generative paradigm, jointly producing sentiment and explanation. To further enhance aspect-oriented reasoning capabilities, we propose a dependency-syntax-guided sentiment cue strategy. This strategy prunes and textualizes the aspect-centered dependency syntax tree, guiding the model to distinguish different sentiment aspects and enhancing its explainability.
To enable explainability, we use MLLMs to construct explanation-augmented datasets for fine-tuning.
Experiments show that our approach not only achieves overall gains in sentiment classification accuracy, but also produces coherent and aspect-grounded explanations.\footnote{https://github.com/ConstMatador/ExplainableMABSA}

\end{abstract}

\section{Introduction}

\begin{figure*}[t]
    \centering
    \vspace{-0.2cm}
    \includegraphics[width=\textwidth, trim=0 12 0 0]{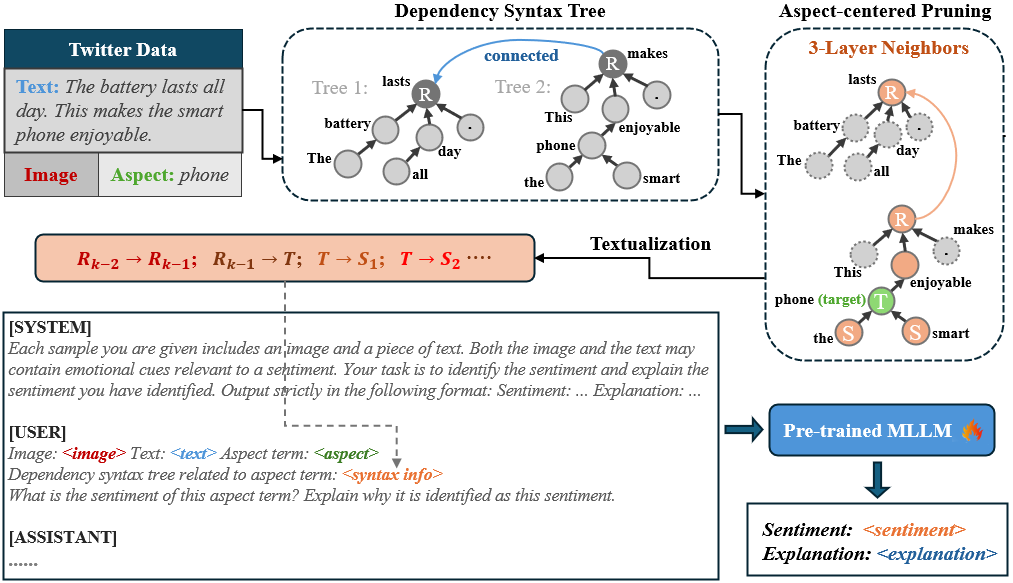}
    \caption{The overall architecture of our approach (the pruned depth shown is three). The process includes three parts: the dependency syntax tree construction phase, the aspect-centered dependency pruning phase, the final textualization and MABSA with dependencies.}
    \label{fig: Overall Architecture}
    \vspace{-0.1cm}
\end{figure*}

%
Aspect-based sentiment analysis (ABSA) aims to identify the sentiment polarity expressed toward specific aspects in text \cite{yan2021unified, xu2021learning, tian2021aspect}. 
With the rapid development of social media, user-generated content increasingly appears in a multimodal form that combines text and images, which has promoted multimodal aspect-based sentiment analysis (MABSA) as an emerging research topic \cite{khan2021exploiting, luo2024panosent, rafiuddin2025adaptisent}. 
By jointly modeling textual and visual information, MABSA enables more fine-grained sentiment understanding that better reflects real-world contexts, and thus plays an important role in practical applications such as social media opinion analysis and brand reputation monitoring \cite{hu2023hierarchical, li2024laldm, yang2025large}.
Furthermore, explainable MABSA is crucial because it can trace the source of sentiment and delve into the reasons for sentiment expression \cite{diwali2023sentiment, maleszka2023survey}.
These explanations not only improve the reliability of sentiment classification but also provide a clear basis for reasoning, making MABSA more suitable for real-world scenarios requiring transparent decision-making.

A growing body of work has made significant contributions to MABSA.
\citet{zhou-etal-2023-aom} proposes AoM to mitigate noise in MABSA via alignment and sentiment aggregation.
\citet{zhu2024joint} enhances JMASA via aspect pre-training and syntactic adaptive learning.
\citet{zhu2025danet} enables fine-grained MABSA via dual-aware alignment and denoising.
The aforementioned studies have yielded promising results, but most existing studies have not delved deeply into the explainable MABSA area.
These approaches merely use MLLMs as feature extractors or discriminative classifiers \cite{cao2024enhanced, xiao2025exploring}.
This fails to fully utilize the autoregressive generative capabilities of pre-trained MLLMs.
Therefore, there is an urgent need to utilize MLLMs for explainable MABSA.
Some studies indicate that text plays a more crucial role in MABSA \cite{yan2021unified, xu2021learning}, and the aspects in the text are the subject of sentiments. A single multimodal social media text usually contains multiple aspects with distinct or even conflicting sentiment polarities, which complicates the identification of sentiment-critical information.
Therefore, it is essential to employ strategies to make MLLMs focus on the sentimental cues within the aspect context.

In this paper, we propose an MLLM-based generative framework for explainable MABSA, which unifies aspect-based sentiment classification and sentiment explanation generation within a single autoregressive process.
To adapt pre-trained MLLMs to this specific task, we construct a training dataset augmented with fine-grained aspect-based sentiment explanations based on existing datasets. 
To enhance the model's ability to understand the context of the target aspect and distinguish sentiments across different aspects, we introduce a dependency-syntax-guided sentiment cue strategy.
Centered on the target aspect, we prune the dependency syntax tree of the text, retaining only the words and syntactic relations most relevant to its sentiment judgment. 
The pruned subtree is then textualized as a prompt for the MLLMs, guiding the model to focus on key clues.
To systematically evaluate the proposed approach and provide a standard benchmark for subsequent study on explainable MABSA, we use a strong MLLM to generate the sentiment explanations, constructing a new dataset with explanations.
Experimental results on Twitter2015 and Twitter2017 show that our proposed approach achieves effective and stable performance improvements on the MABSA benchmark dataset, while providing accurate explanations for the sentiments.

Our main contributions are summarized as follows:
(1) We propose a generative framework based on MLLMs for explainable MABSA that jointly generates sentiment labels and explanations in a single autoregressive process. To evaluate our approach and support future studies on explainable MABSA, we build an explanation-augmented explainable MABSA benchmark on Twitter2015 and Twitter2017.
(2) We introduce a dependency-syntax-guided sentiment cue strategy. This strategy prunes aspect-centered dependency syntax trees and textualizes the subtrees as prompts, to help MLLMs distinguish sentiment cues from different aspects.
(3) Our approach achieves improvements in both sentiment classification accuracy and sentiment explanation quality across multiple model scales and datasets, demonstrating the practical value of our proposed approach.

\section{The Approach}

We propose an MLLM-based generative framework for explainable MABSA that jointly performs aspect-based sentiment classification and explanation generation within a single autoregressive process.
First, we define the problem addressed in our study. Each sample is represented as a 5-tuple $(\mathcal{T}, \mathcal{I}, \mathcal{A}, \mathcal{Y}_S, \mathcal{Y}_E)$, where $\mathcal{T}$ denotes the text, $\mathcal{I}$ denotes the associated image, $\mathcal{A}$ denotes the target aspect, $\mathcal{Y}_S$ denotes the sentiment label, and $\mathcal{Y}_E$ denotes the sentiment explanation.
The sentiment label belongs to a three-class set $\mathcal{Y}_S = \{\textit{negative}, \textit{neutral}, \textit{positive}\}$, and the explanation $\mathcal{Y}_E$ is a natural language sequence.
Given the multimodal input $(\mathcal{T}, \mathcal{I}, \mathcal{A})$, the model generates a structured output containing both the sentiment label and its explanation, which can be formally expressed as: $(\mathcal{Y}_S, \mathcal{Y}_E) = \mathcal{F}(\mathcal{T}, \mathcal{I}, \mathcal{A})$.

In our approach, we use strategies such as fusing dependency syntactic information with pruning to help fine-tune the model.
The overall architecture of our approach is shown in Figure \ref{fig: Overall Architecture}.
Specifically, to enhance aspect-aware reasoning, we incorporate dependency syntax information into the model.
Dependency syntax trees are constructed for individual sentences and connected to form a unified syntactic structure for the entire text.
Aspect-centered pruning is then applied to retain only syntactic components closely related to the target aspect.
The remaining dependency relations are textualized and injected into the model, guiding the model to focus on sentiment-relevant structural cues.
The entire process can be formalized as: 
$(\mathcal{Y}_S, \mathcal{Y}_E) = \mathcal{F}(\mathcal{T}, \mathcal{I}, \mathcal{A}, \mathcal{G})$.

We elaborate on our approach in three sections. 
\ref{sec: Dependency Structure Extraction} demonstrates how we extract dependency syntactic structures from text and form a unified global syntactic representation.
\ref{sec: Aspect-centered Dependency Pruning} proposes an aspect-centered dependency pruning strategy.
\ref{sec: MABSA with Dependencies} textualizes the pruned dependency and uses it as a structured prompt to facilitate explainable MABSA.

\subsection{Dependency Structure Extraction}
\label{sec: Dependency Structure Extraction}

To help MLLMs effectively distinguish the sentiment expressed by different aspects, we construct a unified dependency syntax tree for the text to help understand the contextual sentiment association.

Let the textual input $\mathcal{T}$ be split into $M$ sentences $\{\mathcal{S}_1, \dots, \mathcal{S}_M\}$, where each $\mathcal{S}_m = \{w_{m,1}, \dots, w_{m,n_m}\}$ represents the $m$-th sentence with $n_m$ tokens.
For each sentence $\mathcal{S}_m$, we use an off-the-shelf dependency parser\footnote{E.g., spaCy, Stanford Dependency Parser} to identify token-level directed dependencies and construct a dependency syntax tree.
Formally, the dependency syntax tree of sentence $\mathcal{S}_m$ is represented as a directed graph: $\mathcal{G}_m = (\mathcal{V}_m, \mathcal{E}_m)$, where $\mathcal{V}_m = \{w_{m,1}, \dots, w_{m,n_m}\}$ denotes the set of tokens in sentence $\mathcal{S}_m$, and $\mathcal{E}_m \subseteq \mathcal{V}_m \times \mathcal{V}_m$ denotes the set of directed dependency edges.
$(h_i, w_{m,i}) \in \mathcal{E}_m$ indicates that token $h_i$ serves as the syntactic head of token $w_{m,i}$.
The dependency syntax tree contains a unique root node, denoted as $r_m \in \mathcal{V}_m$, which represents the syntactic root.

To obtain a unified syntactic representation for the entire text, we further introduce inter-sentence structural connections. 
Specifically, for each pair of adjacent sentences $\mathcal{S}_{m-1}$ and $\mathcal{S}_m$ $(m = 2, 3, \dots, M)$, we add a directed edge from the root node $r_{m}$ to $r_{m-1}$.
The set of such inter-sentence edges is defined as:
$\mathcal{E}_m^{\text{inter}} = \{(r_m, r_{m-1}) \mid m = 2, 3, \dots, M\}$.
Finally, the overall syntactic structure of the text $T$ is represented as a unified dependency graph $\mathcal{G} = (\mathcal{V}, \mathcal{E})$, where $\mathcal{V} = \bigcup_{m=1}^{M} \mathcal{V}_m$, and $\mathcal{E} = \bigcup_{m=1}^{M} \mathcal{E}_m \cup \mathcal{E}_m^{\text{inter}}$.

\subsection{Aspect-centered Dependency Pruning}
\label{sec: Aspect-centered Dependency Pruning}

Although the unified dependency tree chain $\mathcal{G} = (\mathcal{V}, \mathcal{E})$
captures rich syntactic relations across the entire text, a large portion of its nodes and edges are irrelevant to the sentiment expressed toward the target aspect.
Moreover, overly complex syntactic structures may introduce noise and hinder effective reasoning in MLLMs.
To address this issue, we perform aspect-centered dependency tree pruning, retaining only the syntactic components that are closely related to the target aspect.

Let $v_\mathcal{A} \in \mathcal{V}$ denote the node corresponding to the target aspect $\mathcal{A}$.
We define two sets of nodes based on their relative positions to $v_\mathcal{A}$ in the dependency tree:
the ancestor nodes (parent direction) and the descendant nodes (child direction).
For a given integer $n$ ($n=\infty$ indicates no pruning), we collect the set of both the ancestor nodes and the descendant nodes within at most $n$ hops from $v_\mathcal{A}$.
Formally, $\text{dist}(u, v)$ denotes the shortest path length between nodes $u$ and $v$ along directed dependency edges.
The retained node set $\mathcal{V}_\mathcal{A}^{(n)}$ is formulated as:
\begin{equation}
\mathcal{V}_\mathcal{A}^{(n)} =
\Big\{
v \in \mathcal{V} \;\Big|\;
\begin{aligned}
& \text{dist}(v, v_\mathcal{A}) \le n \\
\lor\;& \text{dist}(v_\mathcal{A}, v) \le n
\end{aligned}
\Big\}
\end{equation}
Correspondingly, the retained edge set $\mathcal{E}_\mathcal{A}^{(n)}$ includes all edges whose endpoints are both preserved:
\begin{equation}
\mathcal{E}_\mathcal{A}^{(n)} =
\big\{
(u, v) \in \mathcal{E}
\;\big|\;
u \in \mathcal{V}_\mathcal{A}^{(n)} \land v \in \mathcal{V}_\mathcal{A}^{(n)}
\big\}
\end{equation}

The resulting pruned dependency subgraph is defined as: $\mathcal{G}_\mathcal{A}^{(n)}=(\mathcal{V}_\mathcal{A}^{(n)}, \mathcal{E}_\mathcal{A}^{(n)})$. 
This aspect-centered pruned dependency tree focuses on syntactic structures that are most relevant to the sentiment expressed toward the target aspect, while discarding irrelevant or distant information. 
It provides a compact and noise-reduced structural representation for subsequent syntax-aware encoding and generation.

\subsection{MABSA with Dependencies}
\label{sec: MABSA with Dependencies}

To enable the pruned dependency syntax tree to aid in fine-tuning the MLLMs, we textualize it as the input prompt. We consider various textualization approaches, such as CoNLL-U. After empirical comparison, we ultimately choose an edge-based textualization approach.

Given an aspect-centered pruned dependency subgraph \(\mathcal{G}_\mathcal{A}^{(n)} = (\mathcal{V}_\mathcal{A}^{(n)}, \mathcal{E}_\mathcal{A}^{(n)})\), the model converts the dependencies into textual forms. 
Each edge \((u,v) \in \mathcal{E}_\mathcal{A}^{(n)}\) is converted into a text:
\begin{equation}
f_\text{dep}(u,v) = u \xrightarrow{DepRel} v
\end{equation}
where \(u\) and \(v\) denote the parent and child nodes, respectively. \textit{DepRel} denotes the dependency relation between them.  
All edges are sequentially concatenated to obtain the dependency information sequence for the target aspect:
\begin{equation}
\text{DepText}_\mathcal{A} = \bigoplus_{(u,v) \in \mathcal{E}_\mathcal{A}^{(n)}} f_\text{dep}(u,v)
\end{equation}
The dependency information \(\text{DepText}_\mathcal{A}\), together with the guiding instruction $\mathcal{P}$, textual content $\mathcal{T}$, image $\mathcal{I}$, and target aspect $\mathcal{A}$, is provided as input to the MLLMs. 
Conditioned on this input, the model generates an output as:
\begin{equation}
\widehat{\mathcal{Y}}_S, \widehat{\mathcal{Y}}_E = \mathcal{F}_\theta(\mathcal{T}, \mathcal{I}, \mathcal{A}, \text{DepText}_\mathcal{A}, \mathcal{P})
\end{equation}

\section{Experiment Settings}

\subsection{Dataset Construction for Explainable MABSA}

Our experiments are based on two thousand-scale MABSA benchmark datasets, Twitter2015 and Twitter2017 \cite{yu2019adapting}.
The datasets include entries consisting of image, text, target aspect, and sentiment label. 
The statistics of the datasets are shown in Appendix \ref{Apdx:Statistics of Twitter2015 and Twitter2017}.

To enable the fine-tuned model to possess the ability to generate sentiment explanations, we construct explanations for the sentiments with the help of MLLMs.
We use Qwen3-VL-32B \cite{yang2025qwen3}, an MLLM with strong capabilities, to produce explanatory text.
In this process, the model is explicitly provided with the gold standard sentiment label as a constraint during generation.
To further enhance the credibility of the explanations, we randomly sample 10\% of the generated explanations and manually check them in conjunction with the text and image.
%
This inspection confirms that Qwen3-VL-32B can generate accurate, sentiment-consistent explanations when guided by the gold standard labels.
The Twitter datasets enriched with validated explanations are used as our datasets.

\subsection{Baselines}

Our experiments employ Qwen3-VL \cite{yang2025qwen3} and Ministral-3 as backbone models. We compare our proposed pruning variants against these main baseline configurations.

\textbf{Baseline 1: Vanilla MLLMs.} These models are fine-tuned using only the original multimodal inputs (text, image, aspect) without any dependency syntax information. They establish the performance baseline for standard explainable MABSA.

\textbf{Baseline 2: Full-syntax MLLMs.} These models incorporate the complete dependency syntax tree without pruning. They help evaluate the raw contribution of dependency information.

\subsection{Implementation Details}

\begin{table*}[ht]
  \vspace{-0.05cm}
  \centering
  \setlength{\tabcolsep}{1.5mm}
  \scalebox{0.85}{
  \begin{tabular}{l|l|ccccc|ccccc}
    \toprule
    \multirow{3}{*}{\textbf{\vspace{-2ex}MLLM}} &
    \multirow{3}{*}{\makecell[c]{\textbf{\vspace{-2ex}Setting}}} &
    \multicolumn{5}{c|}{\textbf{Twitter2015}} &
    \multicolumn{5}{c}{\textbf{Twitter2017}} \\
    \cmidrule(lr){3-7} \cmidrule(lr){8-12}
    & & 
    \multirow{2}{*}{\makecell[c]{\textbf{BLEU}\\\textbf{-4}}} &
    \multicolumn{3}{c}{\textbf{ROUGE}} &
    \multirow{2}{*}{\makecell[c]{\textbf{BERTScore}\\\textbf{-F1}}} &
    \multirow{2}{*}{\makecell[c]{\textbf{BLEU}\\\textbf{-4}}} &
    \multicolumn{3}{c}{\textbf{ROUGE}} &
    \multirow{2}{*}{\makecell[c]{\textbf{BERTScore}\\\textbf{-F1}}} \\
    \cmidrule(lr){4-6} \cmidrule(lr){9-11}
    & &  
    & \textbf{R-1} & \textbf{R-2} & \textbf{R-L} & 
    & & \textbf{R-1} & \textbf{R-2} & \textbf{R-L} & \\
    \midrule
    \multirow{6}{*}[1.5ex]{\makecell[c]{Qwen3\\-VL-4B}}
      & Vanilla & 33.55 & 60.93 & 37.13 & 49.53 & 92.55 & 34.05 & 61.40 & 37.69 & 49.91 & 92.61 \\
      & Syn (n=$\infty$) & 35.36 & 61.68 & 38.33 & 50.63 & 92.75 & 35.60 & 62.35 & 38.85 & \textbf{51.65} & 92.78 \\
      & Syn (n=1) & 34.95 & 62.01 & 38.37 & 50.69 & 92.76 & 35.51 & \textbf{62.52} & 38.88 & 51.36 & 92.79 \\
      & Syn (n=2) & \textbf{35.45} & \textbf{62.25} & \textbf{38.67} & \textbf{51.06} & \textbf{92.79} & 35.80 & 62.38 & \textbf{39.41} & 51.56 & \textbf{92.81} \\
      & Syn (n=3) & 34.62 & 61.96 & 38.19 & 50.70 & 92.74 & \textbf{36.17} & 62.25 & 39.27 & 51.56 & 92.79 \\
      & CCG & 34.69 & 61.31 & 38.28 & 50.35 & 92.72 & 36.12 & 62.45 & 39.20 & 51.52 & 92.76 \\
    \midrule
    \multirow{6}{*}[1.5ex]{\makecell[c]{Qwen3\\-VL-8B}}
      & Vanilla & 32.63 & 61.98 & 38.34 & 50.60 & 92.73 & 32.98 & 61.86 & 38.71 & 50.75 & 92.72 \\
      & Syn (n=$\infty$) & 34.08 & 62.83 & \textbf{39.79} & 51.86 & 92.94 & 34.48 & 62.85 & 40.59 & \textbf{53.01} & 92.97 \\
      & Syn (n=1) & 34.06 & 62.68 & 39.53 & 51.62 & 92.86 & 34.46 & \textbf{63.31} & 40.02 & 52.37 & \textbf{93.05} \\
      & Syn (n=2) & \textbf{34.28} & \textbf{62.88} & 39.71 & \textbf{52.05} & \textbf{92.95} & \textbf{34.63} & 63.28 & \textbf{40.61} & 52.65 & 92.99 \\
      & Syn (n=3) & 33.85 & 62.43 & 39.27 & 51.48 & 92.86 & 34.56 & 63.08 & 40.25 & 52.32 & 92.98 \\
      & CCG & 33.95 & 62.69 & 39.26 & 51.67 & 92.89 & 34.62 & 62.31 & 39.36 & 51.62 & 92.83 \\
    \midrule
    \multirow{6}{*}[1.5ex]{\makecell[c]{Ministral\\-3-8B}}
      & Vanilla & 23.66 & 54.71 & 31.81 & 43.38 & 90.57 & 24.01 & 55.59 & 33.02 & 44.63 & 90.47 \\
      & Syn (n=$\infty$) & 24.66 & 55.65 & 32.54 & 43.86 & 90.56 & 24.80 & 55.85 & 33.59 & 45.10 & 89.31 \\
      & Syn (n=1) & 24.57 & 55.71 & 32.56 & 44.11 & 90.58 & 24.54 & 55.64 & 33.23 & 44.80 & 90.27 \\
      & Syn (n=2) & \textbf{24.66} & 55.78 & \textbf{32.93} & \textbf{44.55} & 90.38 & \textbf{25.49} & \textbf{56.61} & \textbf{33.99} & \textbf{45.24} & \textbf{90.76} \\
      & Syn (n=3) & 24.48 & \textbf{55.95} & 32.72 & 44.31 & \textbf{90.86} & 24.84 & 55.79 & 33.41 & 44.56 & 90.65 \\
      & CCG & 24.12 & 55.56 & 32.45 & 44.28 & 90.70 & 24.66 & 55.72 & 32.12 & 44.68 & 90.35 \\
    \bottomrule
  \end{tabular}
  }
  \vspace{-0.2cm}
  \caption{Results of generation metrics (\%) in fine-tuning experiments. We perform experiments on three MLLMs, including the vanilla models and multiple variants that integrate dependency syntax with different pruning depths ($n=\infty$ indicates no pruning). Generation metrics include BLEU, ROUGE, and BERTScore.}
  \label{tab: Generation Metrics of Fine-tuned}
\end{table*}

%
Both the baseline and improved models are fine-tuned based on Qwen3-VL-4B, Qwen3-VL-8B, and Ministral-3-8B\footnote{https://docs.mistral.ai/models/ministral-3-8b-25-12}.
Qwen3-VL-4B and Qwen3-VL-8B each contain 36 Transformer layers with hidden state dimensions of 2560 and 4096. Ministral-3-8B consists of 32 layers with a hidden state dimension of 4096.
To achieve efficient parameter fine-tuning, this paper employs Low-Rank Adaptation (LoRA) \cite{hu2022lora} to fine-tune the pre-trained MLLMs. 
The LoRA rank is set to 4, the scaling factor to 16, and the dropout rate to 0.1.
For training settings, the number of epochs is set to 10, and the batch size to 1. The optimizer is AdamW, and the learning rate to 1e-5.

\subsection{Evaluation Metrics}

\begin{table}[t]
  \centering
  \setlength{\tabcolsep}{1.5mm}
  \scalebox{0.85}{
  \begin{tabular}{l|l|cc|cc}
    \toprule
    \multirow{2}{*}{\textbf{MLLM}} & 
    \multirow{2}{*}{\makecell[c]{\textbf{Setting}}} & 
    \multicolumn{2}{c|}{\textbf{Twitter2015}} & 
    \multicolumn{2}{c}{\textbf{Twitter2017}} \\
    \cmidrule(lr){3-4} \cmidrule(lr){5-6}
    & & \textbf{Acc} & \textbf{F1} & \textbf{Acc} & \textbf{F1} \\
    \midrule
    \multirow{6}{*}{\makecell[c]{Qwen3\\-VL-4B}}
      & Vanilla       & 72.8 & 67.7 & 70.4 & 69.5 \\
      & Syn (n=$\infty$) & 74.5 & 70.1 & 71.6 & 71.2 \\
      & Syn (n=1)        & 75.9 & 72.0 & 72.6 & 71.9 \\
      & Syn (n=2)        & \textbf{77.2} & \textbf{72.6} & \textbf{73.0} & \textbf{72.6} \\
      & Syn (n=3)        & 76.0 & 72.1 & 72.0 & 71.2 \\
      & CCG        & 76.3 & 71.5 & 72.5 & 72.4 \\
    \midrule
    \multirow{6}{*}{\makecell[c]{Qwen3\\-VL-8B}}
      & Vanilla       & 73.6 & 70.2 & 71.1 & 71.1 \\
      & Syn (n=$\infty$) & 75.0 & 71.7 & \textbf{73.8} & 72.9 \\ 
      & Syn (n=1)        & 75.6 & 71.4 & 73.0 & 72.0 \\
      & Syn (n=2)        & \textbf{76.7} & \textbf{73.0} & 73.6 & \textbf{73.4} \\
      & Syn (n=3)        & 75.6 & 72.5 & 72.7 & 72.2 \\
      & CCG        & 76.6 & 72.9 & 73.2 & 72.9 \\
    \midrule
    \multirow{6}{*}{\makecell[c]{Ministral\\-3-8B}}
      & Vanilla       & 73.5 & 69.1 & 72.1 & 71.7 \\
      & Syn (n=$\infty$) & 75.5 & 70.0 & 73.2 & 72.3 \\
      & Syn (n=1)        & 75.5 & 70.5 & 71.8 & 71.2 \\
      & Syn (n=2)        & 75.6 & 69.5 & \textbf{74.1} & \textbf{73.3} \\
      & Syn (n=3)        & \textbf{76.6} & \textbf{70.5} & 73.0 & 71.7 \\
      & CCG        & 74.2 & 70.2 & 73.5 & 73.3 \\
    \bottomrule
  \end{tabular}
  }
  \vspace{-0.1cm}
  \caption{Results of classification metrics (\%) in fine-tuning experiments. Classification metrics include accuracy and macro-F1.}
  \label{tab: Classification Metrics of Fine-tuned}
\end{table}

%
We evaluate our approach in terms of both sentiment classification performance and explanation generation quality. 
For sentiment classification, accuracy is used to measure the overall correctness of predictions, while macro-F1 serves as the primary metric to provide a more comprehensive evaluation across different sentiment categories.
For explanation generation, we employ multiple text generation metrics to assess both lexical and semantic aspects. 
BLEU \cite{papineni2002bleu} and ROUGE \cite{lin2004rouge} evaluate the overlap between generated explanations and human references at the n-gram and sequence levels. 
BERTScore \cite{zhang2019bertscore}, computed using the encoder of the RoBERTa-large pre-trained model, measures the semantic similarity between generated and reference texts.

\section{Results and Analyses}

\subsection{Overall Results}

We evaluate the performance of MLLMs under the fine-tuning settings on two tasks: aspect-based sentiment classification and explanation generation.

From the results of aspect-based sentiment classification (Table \ref{tab: Classification Metrics of Fine-tuned}), an overall trend can be observed that incorporating dependency syntactic structures generally improves accuracy and macro-F1 across backbone models and datasets, although the gains vary across pruning depths.
For the Qwen series models, the performance gains brought by incorporating syntactic information are relatively stable across different settings.
In most cases, configurations with moderate pruning depths achieve the best results. 
For Ministral-3-8B, syntax-enhanced models generally achieve better performance than the baseline. 
For the sentiment explanation generation task (Table \ref{tab: Generation Metrics of Fine-tuned}), incorporating syntactic structures usually yields performance improvements in most cases. 
Across BLEU, ROUGE, and BERTScore metrics, syntax-enhanced models mostly outperform or are comparable to the baseline fine-tuned models. 
For the Qwen series models, moderate pruning depths generally lead to better generation performance. 
For Ministral-3-8B, syntactic information generally improves several generation metrics, but the gains are less stable than those of the Qwen-series models.
We also used depccg\footnote{https://github.com/masashi-y/depccg} to assign CCG supertags to the text and used these tags as syntactic features in sentiment analysis, as shown in Table \ref{tab: Generation Metrics of Fine-tuned} and Table \ref{tab: Classification Metrics of Fine-tuned}.

Table \ref{tab: Classification Comparison} shows a comparison of our approach with other approaches on classification metrics. 
Our approach, falling under the category of the prompt-based approaches, achieves competitive results on Twitter2015 and clearly outperforms other prompt-based approaches on Twitter2017.
We also report the results of non-prompt-based approaches which demonstrate higher metrics in Table \ref{tab: Classification Comparison} for reference, mainly due to the introduction of additional information or the adoption of complex model structures.
For instance, GLFFCA and CoolNet rely on sophisticated cross-modal fusion, whereas LRSA and WisDoM introduce auxiliary information such as MLLM-generated rationales or external knowledge.
In contrast, our approach uses only pre-trained MLLMs with syntactic cues, requires no architectural modifications, and jointly optimizes sentiment classification and explanation generation.
These settings somewhat limit the model's ability to specifically optimize classification accuracy.
Under this challenging setting, although our proposed approach does not have an absolute advantage in classification metrics, it still demonstrates good performance and stability.

To quantify the impact of noise when the pruning depth $n$ is large, we conducted experiments on Qwen3-VL-4B on Twitter 2015. 
We replaced the node furthest from the target aspect in the dependency parsing tree with a long and messy string, which simulated and amplified irrelevant noise in the original text. 
The classification accuracy is shown in Appendix \ref{Apdx:Noise Study Results}. The results for Syn (n=$\infty$) and Syn (n=3) showed a certain decrease, but Syn (n=1) and Syn (n=2) were almost unaffected. 
This indicates that when n is relatively large, noise irrelevant to the aspect can affect the analysis results.
Overall, moderate aspect-centered syntactic pruning achieves a good balance between performance improvement and noise suppression. 

\begin{table}[t]
\vspace{-0cm}
\centering
\setlength{\tabcolsep}{1.5mm}
\scalebox{0.8}{
\begin{threeparttable}
\begin{tabular}{l|cc|cc}
\toprule
\multirow{2}{*}{\textbf{Approach}} &
\multicolumn{2}{c|}{\textbf{Twitter2015}} &
\multicolumn{2}{c}{\textbf{Twitter2017}} \\
\cmidrule(lr){2-3} \cmidrule(lr){4-5}
 & \textbf{Acc} & \textbf{F1} & \textbf{Acc} & \textbf{F1} \\
\midrule
\multicolumn{5}{l}{\textit{Non-prompt-based Approaches}} \\
\midrule
SaliencyBERT \cite{wang2021saliencybert} & 77.0 & 72.4 & 69.7 & 67.2 \\
VLP-MABSA \cite{ling2022vision}          & 78.6 & 73.8 & 73.8 & 71.8 \\
CoolNet \cite{xiao2023cross}             & 79.9 & 75.3 & 71.6 & 69.6 \\
LRSA \cite{cao2024enhanced}              & 79.5 & 75.9 & 74.2 & 73.2 \\
WisDoM \cite{wang2024wisdom}             & 81.5 & 78.1 & 77.6 & 76.8 \\
GLFFCA \cite{wang2025aspect}             & 77.7 & 74.2 & 71.2 & 69.5 \\
\midrule
\multicolumn{5}{l}{\textit{Prompt-based Approaches}} \\
\midrule
MultiPoint\tnote{a}\; \cite{yang2023few} & 67.3 & 66.6 & 61.9 & 61.2 \\
Flan-T5-VA \cite{liu2023entity}          & - & 69.7 & - & 68.6 \\
MPFIT \cite{yang2024prompt}              & 76.4 & \textbf{72.6} & 69.0 & 66.7 \\
BIPF \cite{zhu2025bidirectional}         & \textbf{77.3} & 71.2 & 71.1 & 70.4 \\
\midrule
\textbf{Ours}                            & 77.2 & \textbf{72.6} & \textbf{73.0} & \textbf{72.6} \\
\bottomrule
\end{tabular}

\begin{tablenotes}
\footnotesize
\item[a] This approach is specifically designed for few-shot experiments.
\end{tablenotes}
\end{threeparttable}
}
\vspace{-0.05cm}
\caption{Comparison of our approach with other approaches on classification metrics. We categorize these approaches into prompt-based approaches and others.}
\label{tab: Classification Comparison}
\end{table}

\subsection{Zero-shot Results}

\begin{table}[t]
  \centering
  \setlength{\tabcolsep}{1.5mm}
  \scalebox{0.85}{
  \begin{tabular}{l|l|cc|cc}
    \toprule
    \multirow{2}{*}{\textbf{MLLM}} & \multirow{2}{*}{\makecell[c]{\textbf{Setting}}} & 
    \multicolumn{2}{c|}{\textbf{Twitter2015}} & \multicolumn{2}{c}{\textbf{Twitter2017}} \\
    \cmidrule(lr){3-4} \cmidrule(lr){5-6}
    & & \textbf{Acc} & \textbf{F1} & \textbf{Acc} & \textbf{F1} \\
    \midrule
    \multirow{5}{*}{\makecell[c]{Qwen3\\-VL-4B}}
      & Vanilla & 48.1 & 48.5 & 54.9 & 52.3 \\
      & Syn (n=$\infty$) & \textbf{56.9} & \textbf{57.5} & 53.5 & 51.5 \\
      & Syn (n=1) & 55.5 & 56.2 & \textbf{55.1} & \textbf{52.9} \\
      & Syn (n=2) & 55.3 & 56.1 & 53.9 & 51.8 \\
      & Syn (n=3) & 55.0 & 55.7 & 53.6 & 51.6 \\
    \midrule
    \multirow{5}{*}{\makecell[c]{Qwen3\\-VL-8B}} 
      & Vanilla & 59.6 & 59.8 & 59.2 & 59.0 \\
      & Syn (n=$\infty$) & 60.0 & 60.7 & 56.1 & 55.5 \\
      & Syn (n=1) & \textbf{63.9} & \textbf{63.8} & 60.5 & 60.6 \\
      & Syn (n=2) & 62.2 & 62.5 & \textbf{62.7} & \textbf{62.0} \\
      & Syn (n=3) & 62.2 & 61.1 & 58.3 & 57.9 \\
    \midrule
    \multirow{5}{*}{\makecell[c]{Ministral\\-3-8B}} 
      & Vanilla & 33.0 & 54.5 & 34.9 & 55.1 \\
      & Syn (n=$\infty$) & 34.1 & 56.7 & \textbf{35.6} & 55.5 \\
      & Syn (n=1) & 35.4 & 59.2 & 35.5 & 56.5 \\
      & Syn (n=2) & \textbf{35.4} & \textbf{59.2} & 35.5 & \textbf{57.0} \\
      & Syn (n=3) & 34.8 & 57.8 & 35.5 & 56.7 \\
    \bottomrule
  \end{tabular}
  }
  \vspace{-0.2cm}
  \caption{Results of classification metrics (accuracy and macro-F1) (\%) in zero-shot experiments.}
  \label{tab: Classification Metrics of Zero-shot}
\end{table}

%
To evaluate generalization in an unsupervised setting, we assess the models in a zero-shot configuration without fine-tuning, generating predictions directly.
Since there is no fine-tuning to strictly define the output format and length range of the explanations, the generation metrics are low and highly unstable. Therefore, we only focus on the classification metrics.
As shown in Table \ref{tab: Classification Metrics of Zero-shot}, syntactic information can improve zero-shot performance in several settings, but the improvements are less stable than in the fine-tuning setting.
For instance, for Qwen3-VL-8B on Twitter2015, Syn ($n=1$) boosts accuracy and macro-F1 to 63.9\% and 63.8\%, effectively outperforming the baseline.

Notably, the effectiveness of different pruning depths $n$ differs between zero-shot and fine-tuned settings. 
For Ministral-3-8B on Twitter2015, Syn ($n=1,2$) achieves better zero-shot performance, outperforming looser configurations. 
This suggests that, without fine-tuning, overly complex syntactic subgraphs may introduce irrelevant information. 
Stricter local pruning (E.g., $n=1,2$) provides more focused signals and more reliable reasoning.

\subsection{Ablation Study}

In the proposed syntax-aware generative framework, each dependency edge is textualized as a directed relation between two tokens, where the dependency relation explicitly characterizes their syntactic function.
In ablation experiments, we remove dependency relations while keeping the dependency structure and token order unchanged, retaining only the connected token pairs.
The results are reported in Appendix \ref{Apdx:Ablation Study Results}.

From the results, we observe that removing dependency relation labels leads to performance degradation for most models across evaluation metrics. 
This indicates that dependency relation provides additional and effective syntactic cues that help the model more accurately capture aspect-related sentiment expressions. 
Further analysis shows that although the performance drop is relatively moderate, the full model with dependency relation exhibits more stable behavior across different settings. 
This suggests that explicitly encoding syntactic dependency relation enhances the expressiveness of the representation and facilitates more fine-grained aspect-based sentiment modeling.

\subsection{Comparison of Textualization Approaches}

This study considered several other textualization approaches, with the CoNLL-U\footnote{https://universaldependencies.org/format.html} format being the most promising candidate. 
We also conducted fine-tuning experiments using the CoNLL-U format while keeping other settings constant. 
The results are shown in Appendix \ref{Apdx:Results of Fine-Tuning Experiments Using the CoNLL-U Format}.

The results indicate that the CoNLL-U format effectively improves the experimental results compared to the baseline, but is slightly lower than the edge-based textualization approach mentioned above. 
We speculate that this may be because the CoNLL-U format makes the relationships between nodes less intuitive. 
Furthermore, this approach introduces more entries than the edge-based variant, potentially adding noise. 
Nevertheless, both textualization strategies consistently improve performance, demonstrating the robustness and generality of our framework.

\subsection{Intuitive Evaluation of Explanation}

\begin{figure}[t]
    \vspace{-0.05cm}
    \centering
    \includegraphics[width=\linewidth, trim=0 30 0 0]{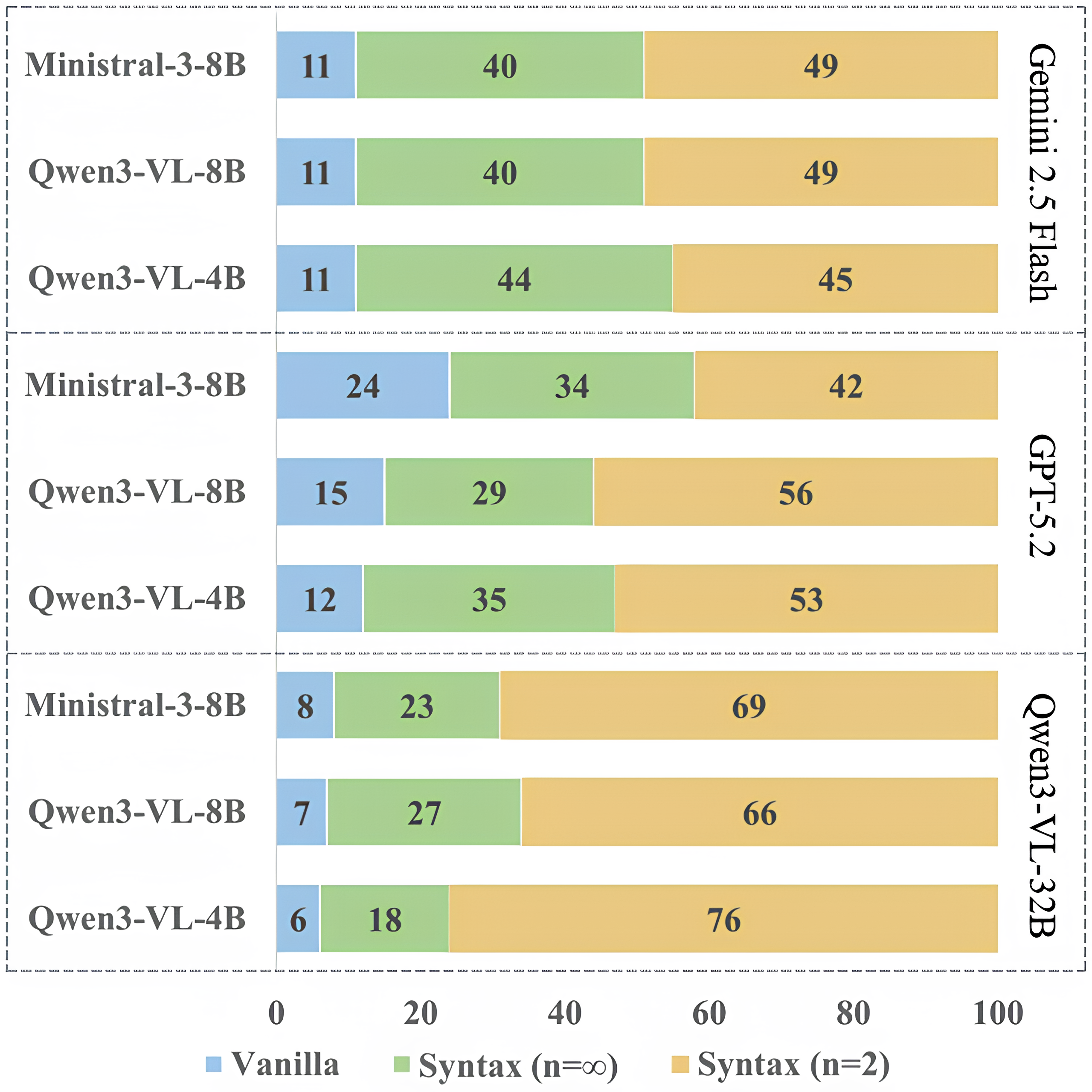}
    \caption{The number of samples (100 in total) that are rated as the best explanation by Gemini 2.5 Flash, GPT-5.2, and Qwen3-VL-32B. Blue, green, and yellow represent results of three experimental groups: Vanilla, Syn ($n=\infty$), and Syn ($n=2$).}
    \vspace{-0.02cm}
    \label{fig: mambo}
\end{figure}

\begin{figure*}[t]
    \vspace{-0.1cm}
    \centering
    \includegraphics[width=0.95\linewidth, trim=0 25 0 0]{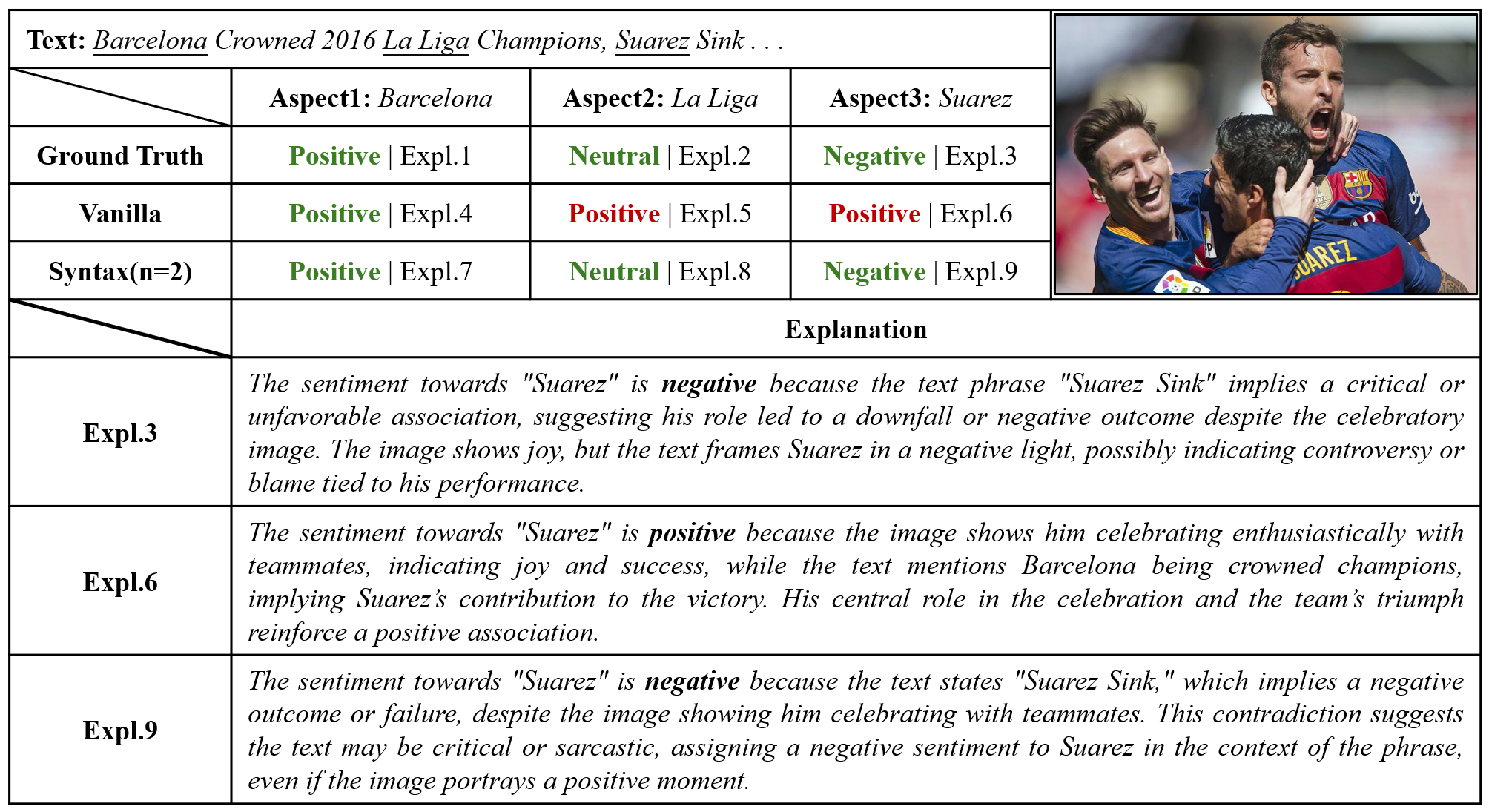}
    \caption{A presentation of sentiment classification results and partial explanations for the case. Our approach successfully distinguished the different sentiments conveyed by different aspects and generated correct explanations.}
    \vspace{-0.05cm}
    \label{fig:case}
\end{figure*}

%
While generation metrics provide a quantitative evaluation, we further assess explanation quality from an intuitive human perspective. 
Specifically, we randomly sample 100 instances from Twitter2015 and compare explanations from three settings: Vanilla, Syn ($n=\infty$), and Syn ($n=2$). We use Gemini 2.5 Flash \footnote{https://docs.cloud.google.com/vertex-ai/generative-ai/docs/models/gemini/2-5-flash?hl=zh-cn}, GPT-5.2 \footnote{https://openai.com/zh-Hans-CN/index/introducing-gpt-5-2}, and Qwen3-VL-32B as the judges to select the best explanation for each sample. We also conduct a small-scale human evaluation by recruiting 10 volunteers, who are asked to choose the best explanation without knowing its source. The results are presented in Appendix \ref{Apdx:Human Evaluation Results}.
The results demonstrate that incorporating dependency syntactic information leads to more intuitive explanations than vanilla MLLMs, while syntactic pruning further enables clearer identification of sentiment cues.

\subsection{Case Study}

As shown in Figure \ref{fig:case}, we present a set of representative samples, where the text-image pair corresponds to three aspects, each expressing a different sentiment. 
The explanation only shows what is necessary and the rest is shown in Appendix \ref{Apdx:All Explanations in the Case Study}. 
The vanilla model fails to clearly distinguish the sentiment of different entities, often mixing the sentiment across aspects. 
For instance, the vanilla model incorrectly predicts a positive sentiment for “Suarez,” whereas our dependency syntax–aware approach captures subject–predicate relations and correctly distinguishes the sentiment of each aspect.
Moreover, our approach provides reasonable explanations for the sentiments. 
The explanation reveals that “Suarez Sink” conveys a negative sentiment despite the celebratory image, indicating that the model correctly prioritizes critical textual cues.

\section{Related Work}

MLLMs integrate visual and textual modalities for cross-modal understanding and generation.
\citet{radford2021learning, alayrac2022flamingo} focus on aligning image–text representations to learn shared semantic spaces, forming the basis for multimodal tasks.
BLIP-2 \cite{li2023blip} bridges vision encoders and LLMs with a lightweight, parameter-efficient Q-Former.
Subsequent models such as MiniGPT-4 \cite{zhu2023minigpt}, InstructBLIP \cite{dai2023instructblip}, and Qwen-VL \cite{wang2024qwen2, yang2025qwen3}, further scale model capacity and improve alignment.

Early studies in aspect-based sentiment analysis explore the integration of textual and visual information.
\citet{ma2018targeted} propose a targeted multimodal framework that aligns text and images to capture visual cues for aspect-specific sentiment prediction.
\citet{xu2019multi} later formalize the MABSA task and introduce the Multi-Interactive Memory Network to model interactions among aspects, text, and images.
Recent studies explore more advanced fusion strategies, including aspect-aware gated fusion to dynamically weight textual and visual features \cite{lawan2025gatemabsa}, joint modeling of emoji, text, and images via mutual attention mechanisms \cite{lou2024emoji}, multi-layer feature fusion with multi-task learning \cite{cai2025multimodal}, and knowledge-enhanced, text-guided frameworks to suppress modality noise \cite{shi2025mmkt}.
These studies merely use MLLMs as feature extractors or discriminative classifiers. 
Unlike them, our study focuses on constructing an explainable MABSA framework that combines sentiment classification and explanation.

\section{Conclusion}

We propose a generative and explainable MABSA framework based on MLLMs that jointly generates sentiment labels and natural language explanations.
A dependency-syntax-guided input structuring strategy enhances aspect-aware reasoning by focusing on aspect-centered syntactic cues.
Experiments on the Twitter2015 and Twitter2017 benchmarks demonstrate consistent improvements in both aspect-based sentiment classification performance and explanation quality, validating the effectiveness and robustness of our approach.

\section*{Limitations}

Our study has some limitations. 
First, dependency parsers perform well on grammatical text but may be less reliable on noisy social media content, potentially affecting the quality of captured sentiment-related dependencies. 
Second, the pruning depth is empirically tuned and dataset-dependent, lacking an automatic mechanism to determine the optimal level, which may cause information loss or redundancy. 
Addressing these issues is an important direction for future studies.

\section*{Acknowledgments}

The work is supported by Zhongguancun Academy (Grant No. c20250509).

\bibliography{custom}

\clearpage
\appendix
\onecolumn

\section{Prompts of Different Tasks}

\begin{table}[H]
\vspace{-0.2pt}
\centering
\renewcommand{\arraystretch}{1.2}
\setlength{\tabcolsep}{8pt}
\begin{tabular}{|p{0.95\textwidth}|}
\hline
\makecell[c]{\textbf{Generating Dataset with Explanations}} \\ \hline
\textbf{[SYSTEM]} \\
Each sample you are given includes an image and a piece of text. Both the image and the text may contain emotional cues relevant to a sentiment label (negative/neutral/positive). \\
Your task is to explain how this sentiment is classified. Your explanation must analyze both the image and the text. Do not rely only on the text. \\
Your response must be a single short paragraph of plain text. Do not include any special formatting, punctuation for lists, markdown symbols, line breaks, or decorative characters. Notice: The explanation should be as brief but accurate as possible. Try to limit it to 50 words or less.\newline
\textbf{[USER]} \\
Image: \textbf{<image>} \;
Text: \textbf{<text>} \;
Aspect term: \textbf{<aspect>} \; \\
Sentiment: \textbf{<sentiment>} \\
Please explain why this sentiment is identified as \textbf{<sentiment>}. \\
\textbf{[ASSISTANT]} \\
···· \\
\hline

\makecell[c]{\textbf{Fine-tuning / Inference of Baseline Model}} \\ \hline
\textbf{[SYSTEM]} \\
Each sample you are given includes an image and a piece of text. Both the image and the text may contain emotional cues relevant to a sentiment (negative/neutral/positive). \\ Your task is to identify the sentiment and explain the sentiment you have identified. \\
Output strictly in the following format: Sentiment: ... Explanation: ... If your output does not follow this format, it is considered incorrect. \newline
\textbf{[USER]} \\
Image: \textbf{<image>} \;
Text: \textbf{<text>} \;
Aspect term: \textbf{<aspect>}. \\
What is the sentiment of this aspect term? Explain why it is identified as this sentiment. \\
\textbf{[ASSISTANT]} \\
···· \\
\hline

\makecell[c]{\textbf{Fine-tuning / Inference of Our Approach}} \\ \hline
\textbf{[SYSTEM]} \\
Each sample you are given includes an image and a piece of text. Both the image and the text may contain emotional cues relevant to a sentiment (negative/neutral/positive). \\ Your task is to identify the sentiment and explain the sentiment you have identified. \\
Output strictly in the following format: Sentiment: ... Explanation: ... If your output does not follow this format, it is considered incorrect. \newline
\textbf{[USER]} \\
Image: \textbf{<image>} \;
Text: \textbf{<text>} \;
Aspect term: \textbf{<aspect>}. \\
Dependency syntax info related to aspect term: \textbf{<syntax info>} \\
What is the sentiment of this aspect term? Explain why it is identified as this sentiment. \\ 
\textbf{[ASSISTANT]} \\
···· \\
\hline

\end{tabular}
\vspace{-0.2pt}
\caption{Prompts of different tasks, including generating dataset with explanations, fine-tuning / inference of baseline and our approach.}
\label{tab:boxed_two_column_prompts}
\end{table}

\clearpage

\section{All Explanations in the Case Study}
\label{Apdx:All Explanations in the Case Study}

\begin{table}[H]
\vspace{-0.2pt}
\centering
\renewcommand{\arraystretch}{1.2}
\begin{tabular}{|>{\centering\arraybackslash}m{1cm}|p{14cm}|}
\hline
\textbf{ID} & \makecell[c]{\textbf{Explanation}} \\
\hline
Expl.1 & \textit{The sentiment towards "Barcelona" is positive because the image shows players in Barcelona jerseys celebrating enthusiastically, with open mouths and joyful expressions, indicating triumph. The text mentions Barcelona being crowned La Liga champions, a clear achievement, reinforcing the positive sentiment associated with the team.} \\
\hline
Expl.2 & \textit{The sentiment towards "La Liga" is neutral because the term is mentioned as the context of the championship, without any evaluative or emotional language attached to it. The image shows celebratory players, but the focus is on Barcelona’s victory, not on La Liga itself. The text references the league factually as the competition being won, not praising or criticizing it.} \\
\hline
Expl.3 & \textit{The sentiment towards "Suarez" is negative because the text phrase "Suarez Sink" implies a critical or unfavorable association, suggesting his role led to a downfall or negative outcome despite the celebratory image. The image shows joy, but the text frames Suarez in a negative light, possibly indicating controversy or blame tied to his performance.} \\
\hline
Expl.4 & \textit{The sentiment towards "Barcelona" is positive because the image shows players celebrating enthusiastically, with visible joy and excitement, indicating a triumphant moment. The text confirms their victory as 2016 La Liga Champions, reinforcing a successful and celebratory context, which aligns with positive sentiment.} \\
\hline
Expl.5 & \textit{The sentiment towards "La Liga" is positive because the text celebrates Barcelona being crowned 2016 La Liga Champions, indicating a successful and triumphant achievement. The image shows players in joyful celebration, reinforcing the positive emotional context associated with winning the league, which directly reflects favorably on La Liga as a prestigious and rewarding competition.} \\
\hline
Expl.6 & \textit{The sentiment towards "Suarez" is positive because the image shows him celebrating enthusiastically with teammates, indicating joy and success, while the text mentions Barcelona being crowned champions, implying Suarez’s contribution to the victory. His central role in the celebration and the team’s triumph reinforce a positive association.} \\
\hline
Expl.7 & \textit{The sentiment towards "Barcelona" is positive because the image shows players in Barcelona jerseys celebrating joyfully, with expressions of triumph and excitement, while the text announces their championship win, directly linking the team to a successful and celebratory outcome.} \\
\hline
Expl.8 & \textit{The sentiment towards "La Liga" is neutral because the term refers to the competition itself, not expressing any positive or negative emotion. The image shows players celebrating, but this reflects team success, not the league’s inherent sentiment. The text mentions Barcelona’s championship, which is a factual outcome, not an evaluation of La Liga as an entity.} \\
\hline
Expl.9 & \textit{The sentiment towards "Suarez" is negative because the text states "Suarez Sink," which implies a negative outcome or failure, despite the image showing him celebrating with teammates. This contradiction suggests the text may be critical or sarcastic, assigning a negative sentiment to Suarez in the context of the phrase, even if the image portrays a positive moment.} \\
\hline
\end{tabular}
\vspace{-0.2pt}
\caption{All explanations in the case study. The explanation results include the ground truth, the vanilla MLLM experimental group, and the experimental group with two pruned layers.}
\end{table}

\clearpage

\section{Statistics of Twitter2015 and Twitter2017}
\label{Apdx:Statistics of Twitter2015 and Twitter2017}

\begin{table}[H]
\vspace{-0.2pt}
\centering
\vspace{-0.2pt}
\setlength{\tabcolsep}{1.5mm}
\scalebox{0.85}{
\begin{tabular}{l|ccc|c}
\toprule
\textbf{Dataset} & \textbf{Train} & \textbf{Dev} & \textbf{Test} & \textbf{Total} \\
\midrule
Twitter2015 & 3,179 & 1,122 & 1,037 & 5,338 \\
Twitter2017 & 3,562 & 1,176 & 1,234 & 5,972 \\
\bottomrule
\end{tabular}
}
\vspace{-0.2pt}
\caption{Statistics of Twitter2015 and Twitter2017.}
\vspace{-0.2pt}
\label{tab:dataset}
\end{table}

\section{Ablation Study Results}
\label{Apdx:Ablation Study Results}

\begin{table}[H]
\vspace{-0.2pt}
  \centering
  \setlength{\tabcolsep}{1.5mm}
  \scalebox{0.85}{
  \begin{tabular}{l|l|cc|cc}
    \toprule
    \multirow{2}{*}{\textbf{MLLM}} & 
    \multirow{2}{*}{\makecell[c]{\textbf{Setting}}} & 
    \multicolumn{2}{c|}{\textbf{Twitter2015}} &
    \multicolumn{2}{c}{\textbf{Twitter2017}} \\
    \cmidrule(lr){3-4} \cmidrule(lr){5-6}
    & & \textbf{Acc} & \textbf{F1} & \textbf{Acc} & \textbf{F1} \\
    \midrule
    \multirow{4}{*}{\makecell[c]{Qwen3\\-VL-4B}}
      & Syn (n=$\infty$) & 72.8 & 67.8 & 70.4 & 70.1 \\
      & Syn (n=1) & 72.6 & 66.8 & 71.2 & 69.8 \\
      & Syn (n=2) & \textbf{73.6} & \textbf{69.0} & \textbf{72.7} & \textbf{72.5} \\
      & Syn (n=3) & 73.5 & 68.0 & 72.1 & 71.3 \\
    \midrule
    
    \multirow{4}{*}{\makecell[c]{Qwen3\\-VL-8B}}
      & Syn (n=$\infty$) & 73.7 & 70.4 & 71.4 & 71.4 \\
      & Syn (n=1) & 74.8 & 71.2 & 72.4 & 72.0 \\
      & Syn (n=2) & \textbf{75.8} & \textbf{72.1} & \textbf{73.0} & \textbf{72.3} \\
      & Syn (n=3) & 75.1 & 72.1 & 72.8 & 72.2 \\
    \midrule
    
    \multirow{4}{*}{\makecell[c]{Ministral\\-3-8B}}
      & Syn (n=$\infty$) & 75.1 & 70.6 & 71.2 & \textbf{72.9} \\
      & Syn (n=1) & 75.7 & \textbf{70.9} & 72.2 & 71.8 \\
      & Syn (n=2) & \textbf{76.3} & 70.5 & \textbf{73.3} & 70.4 \\
      & Syn (n=3) & 75.6 & 70.8 & 70.4 & 69.7 \\
    \bottomrule
  \end{tabular}
  }
  \vspace{-0.2pt}
  \caption{Results of classification metrics (\%) in ablation study.}
  \label{tab: Classification Metrics of Ablation Study}
\end{table}

\begin{table*}[h]
\vspace{-0.2pt}
  \centering
  \setlength{\tabcolsep}{1.5mm}
  \scalebox{0.85}{
  \begin{tabular}{l|l|ccccc|ccccc}
    \toprule
    \multirow{3}{*}{\textbf{\vspace{-2ex}MLLM}} &
    \multirow{3}{*}{\makecell[c]{\textbf{\vspace{-2ex}Setting}}} &
    \multicolumn{5}{c|}{\textbf{Twitter2015}} &
    \multicolumn{5}{c}{\textbf{Twitter2017}} \\
    \cmidrule(lr){3-7} \cmidrule(lr){8-12}
    & & 
    \multirow{2}{*}{\makecell[c]{\textbf{BLEU}\\\textbf{-4}}} &
    \multicolumn{3}{c}{\textbf{ROUGE}} &
    \multirow{2}{*}{\makecell[c]{\textbf{BERTScore}\\\textbf{-F1}}} &
    \multirow{2}{*}{\makecell[c]{\textbf{BLEU}\\\textbf{-4}}} &
    \multicolumn{3}{c}{\textbf{ROUGE}} &
    \multirow{2}{*}{\makecell[c]{\textbf{BERTScore}\\\textbf{-F1}}} \\
    \cmidrule(lr){4-6} \cmidrule(lr){9-11}
    & &  
    & \textbf{R-1} & \textbf{R-2} & \textbf{R-L} & 
    & & \textbf{R-1} & \textbf{R-2} & \textbf{R-L} & \\
    \midrule
    
    \multirow{4}{*}[1.5ex]{\makecell[c]{Qwen3\\-VL-4B}}
      & Syn (n=$\infty$) & 32.52 & 61.02 & 37.25 & 49.62 & 92.57 & 33.46 & 61.55 & 38.35 & 50.54 & 92.63 \\
      & Syn (n=1)        & 31.63 & 60.50 & 36.49 & 48.81 & 92.45 & 33.24 & 61.55 & 38.01 & 50.22 & 92.62 \\
      & Syn (n=2)        & \textbf{32.53} & 60.93 & 37.20 & 49.33 & 92.56 & 33.88 & 61.97 & 38.56 & 50.89 & 92.71 \\
      & Syn (n=3)        & 32.44 & \textbf{61.10} & \textbf{37.25} & \textbf{49.62} & \textbf{92.57} & \textbf{34.10} & \textbf{62.13} & \textbf{39.03} & \textbf{51.17} & \textbf{92.76} \\
    \midrule
    
    \multirow{4}{*}[1.5ex]{\makecell[c]{Qwen3\\-VL-8B}}
      & Syn (n=$\infty$) & 33.39 & 61.76 & 38.06 & 50.31 & 92.69 & 34.82 & 62.39 & 39.63 & 51.53 & 92.82 \\
      & Syn (n=1)        & 34.40 & 62.38 & 38.79 & 51.23 & 92.81 & 35.04 & 62.90 & 40.04 & 51.95 & 92.90 \\
      & Syn (n=2)        & 34.83 & 62.58 & 39.20 & 51.34 & 92.87 & \textbf{35.51} & \textbf{63.06} & \textbf{40.34} & \textbf{52.47} & \textbf{92.95} \\
      & Syn (n=3)        & \textbf{34.84} & \textbf{62.63} & \textbf{39.32} & \textbf{51.63} & \textbf{92.87} & 35.29 & 62.95 & 40.16 & 52.27 & 92.94 \\
    \midrule
    
    \multirow{4}{*}[1.5ex]{\makecell[c]{Ministral\\-3-8B}}
      & Syn (n=$\infty$) & \textbf{24.64} & 55.38 & 32.58 & 44.12 & \textbf{90.84} & 24.61 & 55.20 & 32.82 & 44.25 & 90.35 \\
      & Syn (n=1)        & 24.22 & 55.87 & \textbf{33.23} & \textbf{44.69} & 90.43 & 25.22 & 55.64 & 33.23 & 44.70 & 90.36 \\
      & Syn (n=2)        & 24.45 & \textbf{55.97} & 32.95 & 44.53 & 90.39 & 24.09 & 55.13 & 32.82 & 44.28 & 90.30 \\
      & Syn (n=3)        & 24.02 & 55.03 & 32.21 & 43.67 & 90.71 & \textbf{26.07} & \textbf{56.25} & \textbf{34.21} & \textbf{45.58} & \textbf{90.54} \\
    \bottomrule
  \end{tabular}
  }
  \vspace{-0.2pt}
  \caption{Results of generation metrics (\%) in ablation study.}
  \label{tab: Generation Metrics of Ablation Study}
\end{table*}

\section{Noise Study Results}
\label{Apdx:Noise Study Results}

\begin{table}[H]
\vspace{-0.2pt}
\centering
\vspace{-0.2pt}
\setlength{\tabcolsep}{1.5mm}
\scalebox{0.85}{
\begin{tabular}{l|cccc}
\toprule
\textbf{Setting} & \textbf{Syn ($n=\infty$)} & \textbf{Syn ($n=1$)} & \textbf{Syn ($n=2$)} & \textbf{Syn ($n=3$)} \\
\midrule
Main Experiment & 74.53 & 75.92 & 77.18 & 76.02 \\
Noise Added     & 73.75 & 75.85 & 77.22 & 75.48 \\
\bottomrule
\end{tabular}
}
\vspace{-0.2pt}
\caption{Classification accuracy under the original setting and the noise-added setting.}
\vspace{-0.2pt}
\label{tab:noise_analysis}
\end{table}

\section{Human Evaluation Results}
\label{Apdx:Human Evaluation Results}

\begin{table}[H]
\vspace{-0.2pt}
\centering
\vspace{-0.2pt}
\setlength{\tabcolsep}{1.5mm}
\scalebox{0.85}{
\begin{tabular}{l|ccc}
\toprule
\textbf{Evaluator} & \textbf{Vanilla} & \textbf{Syn ($n=\infty$)} & \textbf{Syn ($n=2$)} \\
\midrule
Tester 1  & 25 & 22 & 53 \\
Tester 2  & 14 & 44 & 42 \\
Tester 3  & 24 & 28 & 48 \\
Tester 4  & 11 & 26 & 63 \\
Tester 5  & 9  & 25 & 66 \\
Tester 6  & 33 & 26 & 41 \\
Tester 7  & 19 & 35 & 46 \\
Tester 8  & 22 & 48 & 30 \\
Tester 9  & 39 & 17 & 44 \\
Tester 10 & 17 & 28 & 55 \\
\midrule
Total & 213 & 299 & 488 \\
\bottomrule
\end{tabular}
}
\vspace{-0.2cm}
\caption{Results of the human evaluation. Each evaluator selected the best explanation among Vanilla, Syn ($n=\infty$), and Syn ($n=2$) for 100 randomly sampled instances.}
\label{tab:human_eval}
\end{table}

\section{Results of Fine-Tuning Experiments Using the CoNLL-U Format}
\label{Apdx:Results of Fine-Tuning Experiments Using the CoNLL-U Format}

\begin{table}[h]
  \vspace{-0.1cm}
  \centering
  \setlength{\tabcolsep}{1.5mm}
  \scalebox{0.85}{
  \begin{tabular}{l|l|cc|cc}
    \toprule
    \multirow{2}{*}{\textbf{MLLM}} & 
    \multirow{2}{*}{\makecell[c]{\textbf{Setting}}} & 
    \multicolumn{2}{c|}{\textbf{Twitter2015}} & 
    \multicolumn{2}{c}{\textbf{Twitter2017}} \\
    \cmidrule(lr){3-4} \cmidrule(lr){5-6}
    & & \textbf{Acc} & \textbf{F1} & \textbf{Acc} & \textbf{F1} \\
    \midrule
    \multirow{4}{*}{\makecell[c]{Qwen3\\-VL-4B}}
      & Syn (n=$\infty$) & 75.1 & \textbf{71.5} & 72.0 & 71.5 \\
      & Syn (n=1) & 75.1 & 69.9 & 72.3 & 71.7 \\
      & Syn (n=2) & 74.7 & 69.8 & 71.9 & 71.3 \\
      & Syn (n=3) & \textbf{75.5} & 70.8 & \textbf{72.8} & \textbf{72.3} \\
    \midrule
    \multirow{4}{*}{\makecell[c]{Qwen3\\-VL-8B}}
      & Syn (n=$\infty$) & 76.0 & 72.7 & 72.9 & 72.5 \\
      & Syn (n=1) & \textbf{76.4} & 72.2 & 72.3 & 71.8 \\
      & Syn (n=2) & 74.9 & 71.1 & 72.8 & 72.3 \\
      & Syn (n=3) & 76.4 & \textbf{72.8} & \textbf{73.5} & \textbf{73.1} \\
    \midrule
    \multirow{4}{*}{\makecell[c]{Ministral\\-3-8B}}
      & Syn (n=$\infty$) & 74.6 & 70.0 & 72.5 & 71.5 \\
      & Syn (n=1) & 75.5 & 70.5 & 73.0 & 72.6 \\
      & Syn (n=2) & \textbf{76.5} & \textbf{70.8} & \textbf{73.3} & \textbf{72.8} \\
      & Syn (n=3) & 75.4 & 70.5 & 73.0 & 72.2 \\
    \bottomrule
  \end{tabular}
  }
  \vspace{-0.2cm}
  \caption{Results of classification metrics (\%) in fine-tuning experiments where CoNLL-U format is adopted.}
  \label{tab: Classification Metrics of CoNLL-U Format Textualization}
\end{table}

\begin{table}[h]
  \vspace{-0cm}
  \centering
  \setlength{\tabcolsep}{1.5mm}
  \scalebox{0.85}{
  \begin{tabular}{l|l|ccccc|ccccc}
    \toprule
    \multirow{3}{*}{\textbf{\vspace{-2ex}MLLM}} &
    \multirow{3}{*}{\makecell[c]{\textbf{\vspace{-2ex}Setting}}} &
    \multicolumn{5}{c|}{\textbf{Twitter2015}} &
    \multicolumn{5}{c}{\textbf{Twitter2017}} \\
    \cmidrule(lr){3-7} \cmidrule(lr){8-12}
    & & 
    \multirow{2}{*}{\makecell[c]{\textbf{BLEU}\\\textbf{-4}}} &
    \multicolumn{3}{c}{\textbf{ROUGE}} &
    \multirow{2}{*}{\makecell[c]{\textbf{BERTScore}\\\textbf{-F1}}} &
    \multirow{2}{*}{\makecell[c]{\textbf{BLEU}\\\textbf{-4}}} &
    \multicolumn{3}{c}{\textbf{ROUGE}} &
    \multirow{2}{*}{\makecell[c]{\textbf{BERTScore}\\\textbf{-F1}}} \\
    \cmidrule(lr){4-6} \cmidrule(lr){9-11}
    & &  
    & \textbf{R-1} & \textbf{R-2} & \textbf{R-L} & 
    & & \textbf{R-1} & \textbf{R-2} & \textbf{R-L} & \\
    \midrule
    
    \multirow{4}{*}[1.5ex]{\makecell[c]{Qwen3\\-VL-4B}}
      & Syn (n=$\infty$) & \textbf{34.00} & 61.98 & \textbf{38.34} & \textbf{50.75} & \textbf{92.76} & 35.01 & 62.59 & 39.65 & 51.80 & 92.82 \\
      & Syn (n=1)        & 33.81 & \textbf{62.00} & 38.26 & 50.72 & 92.75 & 34.50 & 62.29 & 39.25 & 51.54 & 92.80 \\
      & Syn (n=2)        & 33.62 & 61.81 & 38.15 & 50.52 & 92.72 & 34.69 & 62.44 & 39.47 & 51.58 & 92.82 \\
      & Syn (n=3)        & 33.46 & 61.54 & 37.92 & 50.35 & 92.69 & \textbf{35.14} & \textbf{62.79} & \textbf{39.81} & \textbf{51.97} & \textbf{92.87} \\
    \midrule
    
    \multirow{4}{*}[1.5ex]{\makecell[c]{Qwen3\\-VL-8B}}
      & Syn (n=$\infty$) & \textbf{35.76} & 63.22 & \textbf{40.08} & \textbf{52.32} & \textbf{93.01} & \textbf{36.22} & \textbf{63.55} & 41.01 & \textbf{52.85} & \textbf{93.05} \\
      & Syn (n=1)        & 35.70 & \textbf{63.30} & 40.03 & 52.27 & 93.00 & 35.71 & 63.15 & 40.59 & 52.48 & 92.96 \\
      & Syn (n=2)        & 35.02 & 62.83 & 39.36 & 51.65 & 92.89 & 36.00 & 63.29 & 40.81 & 52.73 & 93.01 \\
      & Syn (n=3)        & 35.35 & 63.15 & 39.76 & 52.02 & 92.96 & 36.17 & 63.39 & \textbf{41.04} & 52.78 & 93.02 \\
    \midrule
    
    \multirow{4}{*}[1.5ex]{\makecell[c]{Ministral\\-3-8B}}
      & Syn (n=$\infty$) & 24.61 & 55.49 & 32.57 & 43.73 & 90.74 & 25.78 & 56.00 & 33.44 & 45.22 & 90.02 \\
      & Syn (n=1)        & 25.03 & 55.14 & 32.65 & 44.26 & 90.24 & 25.42 & 56.04 & 33.52 & 44.72 & 90.76 \\
      & Syn (n=2)        & \textbf{25.51} & \textbf{56.17} & \textbf{33.20} & \textbf{44.52} & \textbf{91.28} & \textbf{27.00} & \textbf{56.78} & \textbf{34.50} & \textbf{45.84} & \textbf{90.88} \\
      & Syn (n=3)        & 24.59 & 55.47 & 32.64 & 44.02 & 90.19 & 24.52 & 54.62 & 32.45 & 43.58 & 90.77 \\
    \bottomrule
  \end{tabular}
  }
  \vspace{-0.2cm}
  \caption{Results of generation metrics (\%) in fine-tuning experiments where CoNLL-U format is adopted.}
  \label{tab: Generation Metrics of CoNLL-U Format Textualization}
\end{table}

\end{document}